\documentclass[letterpaper, 10 pt, journal, twoside]{IEEEtran}  %
\usepackage{amsmath,amsfonts,amssymb}
\let\labelindent\relax
\usepackage{enumitem}
\usepackage{prettyref}
\usepackage{mathrsfs}
\usepackage{graphicx}
\usepackage{wrapfig}
\usepackage{subfloat}
\usepackage{MnSymbol}
\usepackage{multirow}
\usepackage{booktabs}
\usepackage[table]{xcolor}
\usepackage{cellspace}
\usepackage{mathtools}
\usepackage{sidecap}
\usepackage[noend]{algpseudocode}
\usepackage{comment}
\usepackage{duckuments}
\usepackage[ruled,vlined,linesnumbered]{algorithm2e}
\usepackage[colorlinks,allcolors=gray,hypertexnames=true]{hyperref}

\algnewcommand{\LineComment}[1]{\State \(\triangleright\) #1}
\algdef{SE}[DOWHILE]{Do}{doWhile}{\algorithmicdo}[1]{\algorithmicwhile\ #1}
\newcommand*{\colorboxed}{}
\def\colorboxed#1#{%
  \colorboxedAux{#1}%
}
\newcommand*{\colorboxedAux}[3]{%
  \begingroup
    \colorlet{cb@saved}{.}%
    \color#1{#2}%
    \boxed{%
      \color{cb@saved}%
      #3%
    }%
  \endgroup
}

\newrefformat{Fig}{Fig.~\ref{#1}}
\newrefformat{fig}{Fig.~\ref{#1}}
\newrefformat{par}{Section~\ref{#1}}
\newrefformat{appen}{Appendix~\ref{#1}}
\newrefformat{sec}{Section~\ref{#1}}
\newrefformat{sub}{Section~\ref{#1}}
\newrefformat{table}{Table~\ref{#1}}
\newrefformat{alg}{Algorithm~\ref{#1}}
\newrefformat{Alg}{Algorithm~\ref{#1}}
\newrefformat{Def}{Definition~\ref{#1}}
\newrefformat{Thm}{Theorem~\ref{#1}}
\newrefformat{Lem}{Lemma~\ref{#1}}
\newrefformat{step}{Step~\ref{#1}}
\newrefformat{ln}{Line~\ref{#1}}
\newrefformat{eq}{Eqn.~\ref{#1}}
\newrefformat{eqn}{Eqn.~\ref{#1}}
\newrefformat{pb}{Problem~\ref{#1}}
\newrefformat{it}{Item~\ref{#1}}
\newrefformat{te}{Term~\ref{#1}}
\def\Eqref Eq:#1:{\eqref{eq:#1}}
\newrefformat{Eq}{Equation~\Eqref#1:}

\newcommand{\E}[1]{\mathbf{#1}}
\newcommand{\TE}[1]{\textbf{#1}}

\newcommand{\FPPROW}[2]{{\partial{#1}}/{\partial{#2}}}

\newcommand{\FPP}[2]{\frac{\partial{#1}}{\partial{#2}}}

\newcommand{\FDD}[2]{\frac{d{#1}}{d{#2}}}

\newcommand{\argmin}[1]{\underset{#1}{\E{argmin}}\;}
\newcommand{\argminP}[1]{\E{argmin}\;}

\newcommand{\argmaxP}[1]{\E{argmax}\;}
\newcommand{\ST}{\E{s.t.}\;}

\definecolor{darkgreen}{HTML}{186a3b}

\usepackage[normalem]{ulem}
\usepackage{xcolor}
\definecolor{Blue}{rgb}{1,0,0}
\definecolor{Orange}{rgb}{0.929,0.49,0.192}

\newcommand{\revise}[1]{\textcolor{black}{#1}}

\usepackage{placeins}

\begin{document}

\title{\Large\bf One-Shot Real-to-Sim via End-to-End Differentiable Simulation and Rendering}

\author{Yifan Zhu$^{1}$, Tianyi Xiang$^{1}$, Aaron M. Dollar$^{1}$, and Zherong Pan$^{2}$
\thanks{Manuscript received: December 8, 2024; Revised: March 7, 2025; Accepted: April 7, 2025.}
\thanks{This paper was recommended for publication by Editor Lucia Pallottino upon evaluation of the Associate Editor and Reviewers’ comments.}
\thanks{$^{1}$Y. Zhu, T. Xiang, and A. Dollar are with the Department of Mechanical Engineering and Materials Science, Yale University, New Haven, United States. {\tt\small \{yifan.zhu, tianyi.xiang, aaron.dollar\}@yale.edu} $^{2}$Z. Pan is an independent researcher. {\tt\small zherong.pan.usa@gmail.com}}
\thanks{Digital Object Identifier (DOI): see top of this page.}
}

\markboth{IEEE ROBOTICS AND AUTOMATION LETTERS. PREPRINT VERSION. April, 2025}
{Zhu \MakeLowercase{\textit{et al.}}: One-Shot Real-to-Sim via End-to-End Differentiable Simulation and Rendering}

\maketitle

\begin{abstract}
Identifying predictive world models for robots from sparse online observations is essential for robot task planning and execution in novel environments. However, existing methods that leverage differentiable programming to identify world models are incapable of jointly optimizing the geometry, appearance, and physical properties of the scene. In this work, we introduce a novel \revise{rigid} object representation that allows the joint identification of these properties. Our method employs a novel differentiable point-based geometry representation coupled with a grid-based appearance field, which allows differentiable object collision detection and rendering. Combined with a differentiable physical simulator, we achieve end-to-end optimization of world models or rigid objects, given the sparse visual and tactile observations of a physical motion sequence. Through a series of \revise{world model identification} tasks in simulated and real environments, we show that our method can learn both simulation- and rendering-ready rigid world models from only one robot action sequence. The code and additional videos are available at our project website: \url{https://tianyi20.github.io/rigid-world-model.github.io/}.
\end{abstract}

\section{Introduction}
\revise{ An accurate internal model of a robot about how its actions can affect the surrounding environment is essential for robot planning and control. Such a model, which we refer to as a \textit{world model}, needs to render realistic raw observations such as RGB images from arbitrary viewpoints and predict consistent and accurate physical interactions. However, constructing such a model from raw observations in novel real-world environments remains challenging as it requires the identification of the geometry parameters that describe the shape of all objects (e.g. vertices and faces of a mesh), appearance parameters that define how the objects look when rendered (e.g. color and reflectance), and physical parameters (e.g. mass) of the objects in the scene. These parameters are usually partially observable, and robots are typically limited in time and computational resources.}


Recently, there has been growing interest in learning world models from large offline datasets of action-labeled videos using generative modeling techniques~\cite{ha2018world,Bruce2024world,Zhu2024world,Yang2023world}. \revise{However, these black-box models are susceptible to distribution shifts and cannot infer properties such as the coefficient of friction. In addition, they are not physically consistent and cannot provide physical information such as contact forces, which are essential for downstream tasks. Meanwhile, an alternative approach that identifies the geometry, appearance, and physical parameters (GAP) of the environment with strong priors coming from knowledge of physics can result in a generalizable and physically consistent world model.}


Many existing works employ differentiable simulators as strong physics priors that allow efficient identification of physical parameters such as inertia and coefficient of friction~\cite{Song-RSS-20,9363565, ehsani2020use}. Differentiable simulators allow the gradient-based optimization of mass-inertial properties and frictional coefficients to match a physical motion sequence to sparse robot observations. However, these works assume known geometries and appearances of the objects in the scene and do not allow the algorithm to adapt the GAP simultaneously.

On the other hand, we have witnessed recent advances in learnable geometry and appearance models, such as Neural Radiance Fields (NeRF)~\cite{mildenhall2021nerf} and Gaussian Splatting (GS)~\cite{kerbl20233d}. \revise{These methods build the rendering equation into a learnable representation to enable the identification of geometries and appearances from raw observations.} However, rigid body simulators~\cite{erez2015simulation} typically require the use of volumetric representations with a clear definition of object surfaces such as convex hulls to detect collisions and penetration depths. \revise{Unfortunately, NeRF and GS are incompatible with the requirements of rigid body simulators since NeRF represents objects with a continuous neural field and GS with individual 3D Gaussians.} To the best of the authors' knowledge, no existing method allows the simultaneous identification of the GAP properties of a world model of rigid objects from sparse robot observations.



\begin{figure}
    \centering
    \includegraphics[trim=3.1cm 4.2cm 6.5cm 3.2cm,clip,width=1\linewidth]{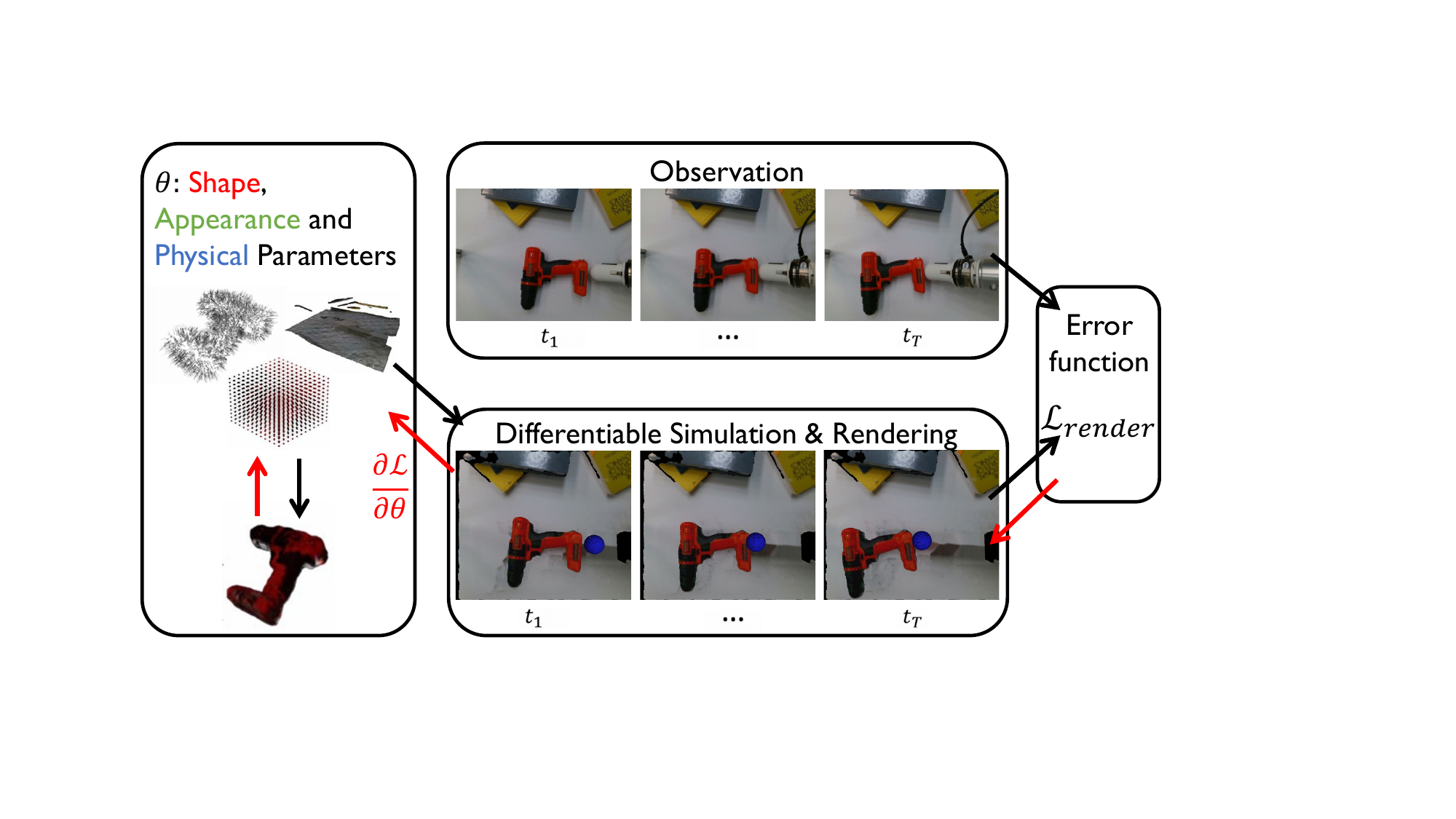}
    \caption{\revise{From the visual and tactile observations of a single robot push (top), our method jointly optimizes the shape, appearance, and physical parameters of a world model consisting of rigid objects in the form of a rigid body simulator (bottom, the robot arm is not rendered in this picture and the end-effector is treated as a floating blue sphere robot).}}
    \label{fig:spotlight}
\end{figure}

To address these challenges, this work presents a rigid object representation that is compatible with general-purpose rigid body simulators and allows the joint optimization of GAP. As shown in~\prettyref{fig:spotlight}, based on this representation, our work enables the identification of a \revise{rigid} world model in the form of a full-fledged rigid body simulator from the observations of one robot push. Our proposed representation is the combination of a recently proposed point-based shape representation Shape-as-Points (SaP)~\cite{peng2021sap} and a grid-based appearance field. SaP parameterizes an object's geometry and topology using a set of surface points along with normal directions. It then uses differentiable Poisson reconstruction to recover a smooth indicator field of object occupancy, which can be converted to a mesh using a differentiable marching cubes algorithm~\cite{liao2018deep}. The texture of the vertices of the mesh is then obtained by interpolating the appearance grid. Employing the mesh in a differentiable rigid body simulator~\cite{strecke2021_diffsdfsim} that provides gradients for the physical parameters and contact points of the objects, our method constructs a fully differentiable pipeline for jointly optimizing the GAP. Our contributions are:

\begin{itemize}
\item A jointly differentiable representation of the shape, appearance, and physical properties of rigid objects.
\item An algorithm for identifying world models online from sparse robot observations, which we refer to as real-to-sim, with an end-to-end differentiable simulation and rendering pipeline.
\end{itemize}

We evaluate our method on identification problems in both simulated and real-world environments. The results show that our method can infer accurate world models from a single episode of robot interactions with the environment. 


\section{Related Work}
Our work is closely related to differentiable rigid body simulators, learnable geometry and appearance models, and identifying world models, and we review these areas of study in this section.

\subsection{Differentiable Rigid Body Simulator}
Rigid body simulators are essential tools in robotics and engineering for testing, verification, perception, control, and planning. Traditional rigid body simulators are not differentiable, but there have been many recently proposed differentiable rigid body simulators~\cite{strecke2021_diffsdfsim,Werling-RSS-21,Xu-RSS-21,NEURIPS2021_8e296a06,howell2022dojo, geilinger2020add} for facilitating downstream system identification, robot planning, and policy optimization tasks. Different strategies are adopted to enable the calculation of gradients for the underlying non-differentiable contact dynamics, including employing a smooth contact model~\cite{geilinger2020add, Xu-RSS-21,NEURIPS2021_8e296a06}, using sub-gradients of the linear complementarity problem~\cite{Werling-RSS-21}, and implicit gradients of nonlinear optimization~\cite{strecke2021_diffsdfsim,howell2022dojo}. However, most of these methods do not provide gradients with respect to the geometry, with the exception of ~\cite{howell2022dojo,strecke2021_diffsdfsim,Xu-RSS-21}. In this work, we adopt the simulator proposed by Strecke et al.~\cite{strecke2021_diffsdfsim} for its physical realism, numerical stability, and fast computation from GPU acceleration.

\subsection{Learnable Geometry and Appearance Models}
Learning 3D geometry and appearance models from 2D raw images is vital to robots' understanding of the physical world. Earlier research has focused on learning only the 3D geometries without appearance, including point-cloud-based~\cite{achlioptas2018learning} models, convex-hull-based~\cite{deng2020cvxnet} models, and learning implicit signed distance functions~\cite{pfrommer2021contactnets}. Neural radiance fields (NeRF) is the first method that enables a continuously learnable model for the full 3D appearance of objects and scenes, which uses neural networks to parameterize the spatial appearance properties and implicitly learn the 3D geometry. More recently, Kerbl et al. proposed Gaussian Splatting (GS)~\cite{kerbl20233d}, a non-parametric method that represents the appearance of the scene with 3D Gaussians and significantly improves the training and rendering speeds due to their fit for fast GPU/CUDA-based rasterization. These algorithms are capable of learning detailed 3D object and scene appearances from sparse image-based observations. However, NeRF and GS lack a clear definition of rigid object surfaces as NeRF represents objects with a neural field and GS with individual 3D Gaussians. Therefore, while there are some initial attempts at integrating them with rigid body simulators~\cite{sharp2022spelunking,xie2024physgaussian}, research for robust, physically correct, and differentiable collision detections with these models is still ongoing. In addition, NeRF and GS require many diverse views of an object, which is unrealistic in typical robotic manipulation applications.

Compared to standard 3D representations such as point clouds, which do not allow volumetric collision detection, or meshes, which do not allow large geometric and topological changes during optimization, our SaP-based methods enjoy the best of both worlds. Combined with a differentiable renderer, our object representation then achieves end-to-end image-based shape optimization.

\subsection{World Models}
\revise{Traditional system identification methods~\cite{aastrom1971system} identify only the dynamics parameters from full state information.} However, to support diverse downstream robot tasks in the real world, world models need to be built from raw observations and support both accurate dynamics prediction and photorealistic novel view synthesis. While existing works have identified world models from raw image observations using differentiable simulators~\cite{howell2022dojo,strecke2021_diffsdfsim,Xu-RSS-21}, none supports simultaneous optimization of GAP. Another line of work closely related to ours is image-based generative world modeling. These works aim to predict the next RGB frame based on the current frame and action. These models are learned by training on diverse datasets with generative modeling techniques such as variational autoencoders~\cite{ha2018world,Bruce2024world} and diffusion~\cite{Zhu2024world, Yang2023world}. The key differences between our method and these works are that our simulation, grounded in physics, is always physically consistent and is a general-purpose \revise{rigid} simulator that can provide physical information such as contact forces. Purely data-driven world models generalize poorly to novel scenarios and their lack of physical information severely limits their application to downstream robot tasks. \revise{Finally, a recent work~\cite{abou-chakra2024physically} proposed a method to use Gaussian Splatting along with a particle-based simulator to track and reconstruct a moving scene. Instead of identifying the physical parameters of the scene, the method optimizes virtual forces attached to each particle such that they match the observed object trajectory. Therefore, although the method can be used as a world model for prediction, the accuracy is severely limited. We include this method as a baseline in our experiments in Sec.~\ref{sec:result} and demonstrate the limitation of this method.}

\begin{figure*}[t!]
\centering
    \includegraphics[trim=0cm 6.5cm 5cm 0cm,clip,width=0.8\linewidth]{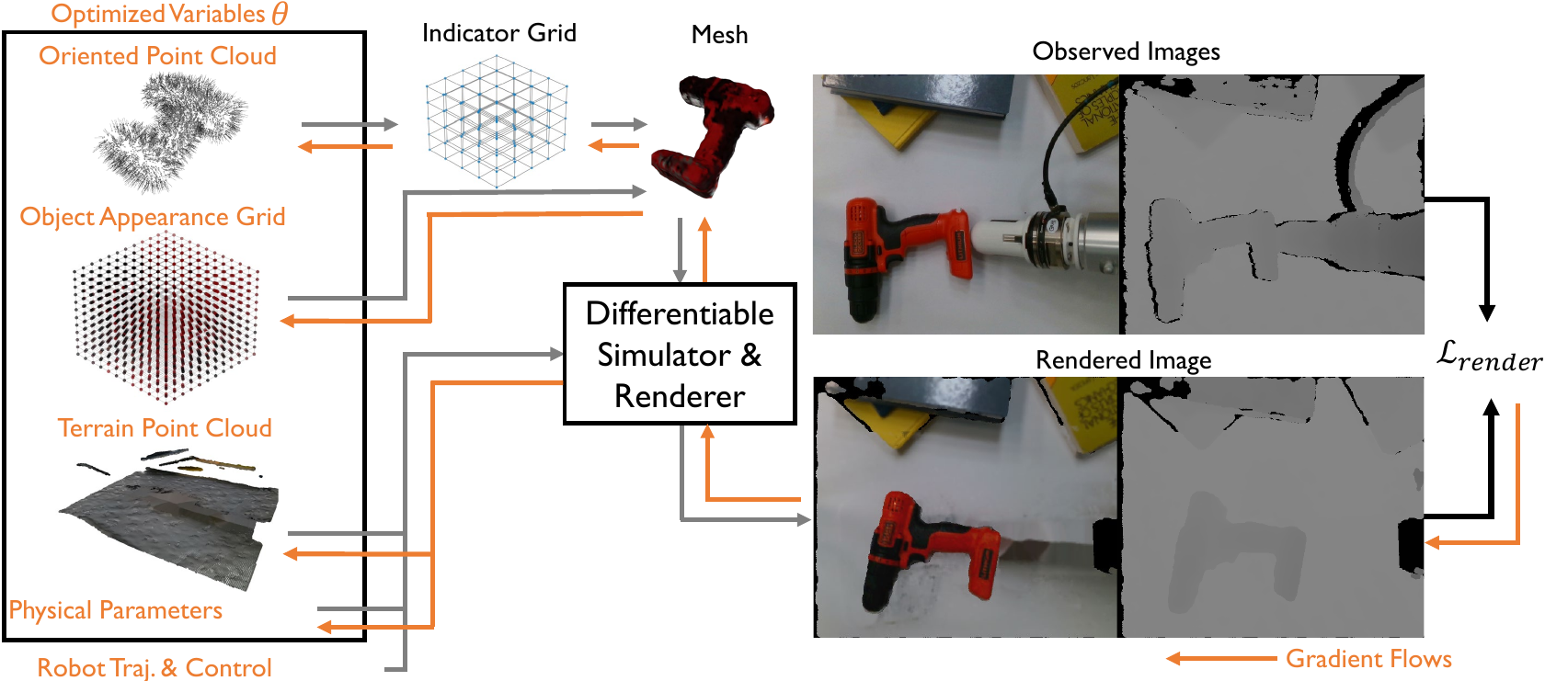}
    \put(-332,159){\footnotesize \textcolor{Orange}{$\mathcal{P}$}}
    \put(-330,118){\footnotesize \textcolor{Orange}{$\mathcal{\psi}$}}
    \put(-333,63){\footnotesize \textcolor{Orange}{$\mathcal{P}_t$}}
    \put(-345,15.5){\footnotesize \textcolor{Orange}{$M,\mu$}}
    \put(-333,1){\footnotesize \textcolor{Orange}{$\langle e^t,u^t \rangle$}}
    \put(-255,166.5){\footnotesize $\chi$}
    \caption{Overview of the proposed fully differentiable pipeline for world model identification from sparse robot observations. Our object representation couples an oriented point cloud $\mathcal{P}$ and a 3D appearance grid $\mathcal{\psi}$. Through a differentiable Poisson solver and differentiable marching cubes, the oriented point cloud is converted to an indicator grid $\chi$ and then a mesh, whose vertex textures are interpolated from the appearance grid $\mathcal{\psi}$. Feeding the object mesh, physical parameters $M$ and $\mu$, the terrain point cloud $\mathcal{P}_t$, and the robot pushing trajectory and control $\langle e^t,u^t \rangle$ into a differentiable rigid body simulator and renderer, the predicted scenes can be rendered. Calculating the loss against observed RGB-D images, the scene shape, appearance, and physical parameters are jointly optimized with gradient descent.}
    \label{fig:method}
\end{figure*}
\section{Problem Definition}

In this section, we describe our formulation of world model identification. We assume the environment consists of a rigid object and rigid terrain, whose physical properties and appearances are parameterized by $\theta$. A robot, equipped with joint encoders and end-effector force sensors, interacts with the object at $T$ time instances: $t_1,\cdots,t_T$, with a fixed time step $\delta t$. At each time step, the robot observes its end-effector pose $e^t\in\text{SE}(3)$ and contact force $f^t\in\mathbb{R}^3$. Further, the robot is equipped with an RGB-D camera with known intrinsics that observes the object through image $o^t\in\mathbb{R}^{H\times W\times 4}$ at camera pose $c^t\in\text{SE}(3)$. We further assume an image segmentation mask $m^t\in\mathbb{R}^{H\times W\times 4}$ is provided for the object, robot, and terrain. Therefore, the robot observations are a sequence $\mathcal{O}=\{\langle t,e^t,f^t,o^t,c^t,m^t\rangle\}$, and our goal is to estimate $\theta$ from the set of sparse observations $\mathcal{O}$. We formulate this problem as a physics-constrained optimization by introducing a full-fledged physics simulator function $q^{i+1}, \dot{q}^{i+1}=g(q^i,\dot{q}^{i},u^i,\theta)$ that can differentiate through objects' appearance, geometry, and physical parameters. Here, $q^t$ and $\dot{q}^{i+1}$ are the object and robot end-effector poses and velocities at timestep $i$ and $u^i$ is the applied robot force at the end-effector, which is equal in magnitude to the sensed contact force but opposite in direction.


Given such a simulator, the world model identification problem is formulated as solving the following optimization:
\begin{equation}
\label{eq:SI}
\begin{aligned}
\argmin{\theta} & \sum_{t=t_1}^{t_T}\mathcal{L}\Big( \hat{o}^t \big(q^t(\theta), \theta\big),o^t \Big) \\
\ST\quad & q^{i+1},\dot{q}^{i+1}=g(q^i,\dot{q}^{i},u^i,\theta)\quad\forall t=1,\cdots,T-1.
\end{aligned}
\end{equation}
The optimization is solved over a physical motion sequence of $T$ timesteps, with the objective function $\mathcal{L}$ encourages the simulated observation $\hat{o}^t\big(q^t(\theta), \theta\big)$ to match the ground-truth observation $o^t$. 



\section{Method}
In this section, we first detail the object representation, which is key to our method. Then we describe the differentiable simulator and details on solving the optimization described in Eqn.~\ref{eq:SI}.

\subsection{Differentiable Object Representation}
An ideal object representation for world model identification needs to be flexible to allow learning of complex object geometries, topologies, and appearance properties while being compatible with rigid body simulators for collision detection. Topology-agnostic geometries such as point clouds~\cite{achlioptas2018learning} and GS~\cite{kerbl20233d} do not allow one to calculate the penetration depth between bodies. On the other hand, meshes~\cite{Xu-RSS-21} do not allow large geometric and topological changes. 

We find that the SaP framework~\cite{peng2021sap}, when augmented by additional appearance properties poses an ideal representation for our purpose. Briefly, this framework represents the object using a point cloud with normals on the object surface, denoted as $\mathcal{P}=\{( p\in\mathbb{R}^3,n\in\mathbb{R}^3)\}$. These normal directions induce a discrete vector field $v(x)=\sum_{\langle p,n\rangle\in\mathcal{P}}n\mathbb{I}[x=p]$. SaP then uses Poisson reconstruction~\cite{kazhdan2006poisson} to recover an underlying implicit indicator field $\chi(x)$ that describes the occupancy of the solid geometry, i.e. whether $x$ is inside or outside the geometry, and matches its gradient field with $v(x)$ by solving the variational problem:
\begin{align*}
\argmin{\chi}\int_\Omega\|\nabla\chi(x)-v(x)\|^2,
\end{align*}
which amounts to solving the Poisson equation $\Delta\chi=\nabla\dot v$. SaP discretizes the indicator field $\chi$ on a uniform grid domain $\Omega$, which allows the efficient solution of $\chi$ via GPU-accelerated Fast Fourier Transform (FFT) with well-defined derivatives. We use a $128\times 128\times 128$ discretized grid $\chi$ for all the experiments in this paper.

The indicator field $\chi$ is then transformed to a triangle mesh $\mathcal{M}$ with a differentiable marching cubes algorithm~\cite{liao2018deep}. Collision detection can then be easily achieved with standard techniques for meshes. To enable appearance modeling, we further augment with a grid of appearance properties, with the same grid resolution as the one storing the indicator field $\chi$. The appearance property is then propagated to the mesh vertices via tri-linear interpolation. In this work, we only store and render the color field, denoted as $\psi$, but other appearance properties can be incorporated in the same manner as required by more advanced differentiable rendering equations. The mesh can then be rendered using any differentiable renderer framework such as~\cite{liu2019soft,Laine2020diffrast}, for which we use the open source implementation in PyTorch3D~\cite{ravi2020pytorch3d}. Specifically, at the time instance $t$, we invoke the renderer with the object transformed to $q^t$ and the camera transformed to $c^t$. Our parametrization of the object's physics and appearance is defined as:
\begin{align*}
\theta\triangleq\langle M(q^i),\mu,\mathcal{P},\psi \rangle,
\end{align*}
where the first two parameters are mass-inertial properties and frictional coefficients, and the last two parameters are the oriented point cloud for SaP and color field.

For the terrain, we simply use an oriented and colored point cloud $\mathcal{P}_t$ to represent the terrain as we do not need to simulate interactions between 2 terrains. The terrain is rendered from the colored point cloud with an alpha compositor~\cite{wiles2020synsin} also using the PyTorch3D library and we set the radius of each point to be 0.015\,m.

\subsection{Differentiable Simulator}
For our application, we only consider unconstrained rigid body dynamics with dry frictional contacts. Note that additional physical constraints for describing objects such as soft bodies and articulated objects can be potentially incorporated into our framework and its differentiation has been well-studied, e.g. in~\cite{NEURIPS2021_8e296a06}. 

The governing equation of motion for rigid bodies and the time discretization method are well-established, and we refer the readers to Anitescu et al.~\cite{anitescu1997formulating} for details. The equation is summarized as follows:
\begin{align}
\label{eq:EOMA}
M(q^i)\ddot{q}^i=C(q^i,\dot{q}^i)+J^iu^i+J^\perp\tau^\perp+J^\parallel\tau^\parallel,
\end{align}
with $M(q^i)$ being the generalized mass matrix, $C(q^i,\dot{q}^i)$ being the centrifugal, Coriolis, and gravitational force, $J^i,J^\perp,J^\parallel$ being the Jacobian matrix for the external, normal, and tangent contact forces at all the detected contact points, respectively. Finally, $\tau^\perp,\tau^\parallel$ are the contact forces. At each time step, a mixed linear complementarity problem (LCP) is solved to calculate the constraint forces $\tau^\perp,\tau^\parallel$, yielding the final acceleration $\ddot{q}^i$, and we then integrate the configuration forward in time~\cite{cline2002rigid,stewart1996implicit} as: 
\begin{align}
\label{eq:EOMB}
\dot{q}^{i+1}=\dot{q}^i+\ddot{q}^i\delta t\quad
q^{i+1}=q^i+\dot{q}^i\delta t,
\end{align}
with $\delta t$ being the timestep size. The mixed LCP problem is formulated as:
\begin{align}
\label{eq:EOMC}
\begin{cases}
0\leq\tau^\perp
&\perp J^{\perp T}\dot{q}^{i+1}\geq0\\
0\leq\tau^\parallel
&\perp \lambda e+J^{\parallel T}\dot{q}^{i+1}\geq0\\
0\leq\lambda
&\perp \mu\tau^\perp-e^T\tau^\parallel\geq0,
\end{cases}
\end{align}
with $e$ being the unit vector and $\mu$ being the frictional coefficient. $\lambda$ is an auxiliary variable encoding the stick or slip frictional state. To differentiate through the simulator, we adopt the differentiation technique proposed by~\cite{Werling-RSS-21,strecke2021_diffsdfsim}, where the result of the LCP is made differentiable by solving with a primal-dual method and performing sensitivity analysis at the solution to yield derivatives with respect to the problem data. In this way, the derivatives propagate the gradient information to the Jacobian matrix $J^{\perp,\parallel}$, and finally to the object geometric parameters $\mathcal{P}$. In summary, \prettyref{eq:EOMA},\ref{eq:EOMB},\ref{eq:EOMC} defines our differentiable simulator function $g$. In particular, we adopt the differentiable simulator proposed by Strecke et al.~\cite{strecke2021_diffsdfsim} for its fast implementation on GPU.

\begin{figure}[t!]
\centering
    \includegraphics[trim=0cm 17cm 31cm 0cm,clip,width=1\linewidth]{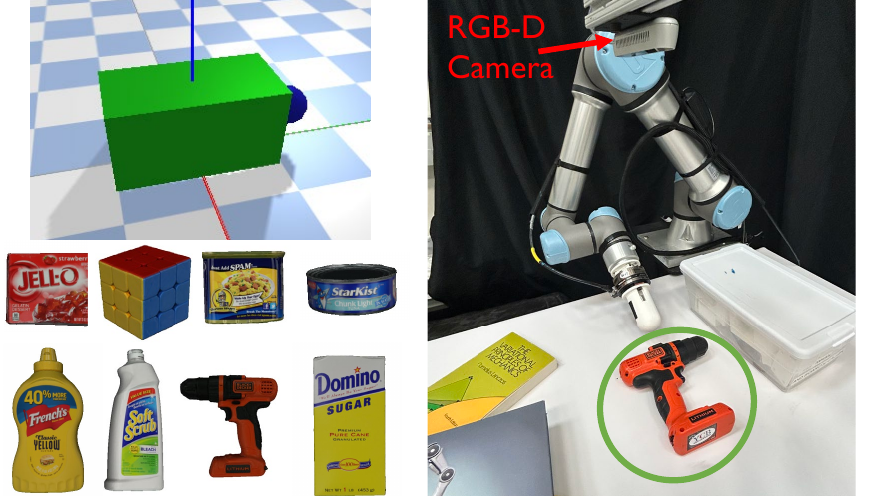}
    \put(-213,-8){\small Simulation }
    \put(-90,-8){\small Real World}
    \caption{\revise{The experiment setups for the simulation (left) and physical (right) experiments. 9 objects are used for simulation with the PyBullet simulator, including 8 YCB objects and a green box. For the real-world experiments, three YCB objects (\texttt{Drill}, \texttt{Mustard}, and \texttt{Sugar}) are used. A UR5e arm equipped with a pusher and a ATI Gamma F/T sensor and an overhead Realsense D435 RGB-D camera are used. Note that only the circled object in the real-world setup is the object of interest and everything else is treated as the static terrain.}}
    \label{fig:setups}
\end{figure}

\subsection{World Model Identification}
Even with our jointly differentiable physical and appearance models, solving~\prettyref{eq:SI} can still be rather challenging. This is mainly because our initial guess can be very poor, especially in the occluded region. As a result, the na\"ive gradient descent method can take many iterations and is prone to converging to poor local minima. To mitigate this, we use two stages of optimization, and we further leverage 3D foundation models trained on web-scale data to generate reasonable initial guesses of the rigid object in the scene from partial visual observations.

\begin{figure}[t!]
\centering
    \includegraphics[trim=0cm 12.3cm 1.5cm 0cm,clip,width=1\linewidth]{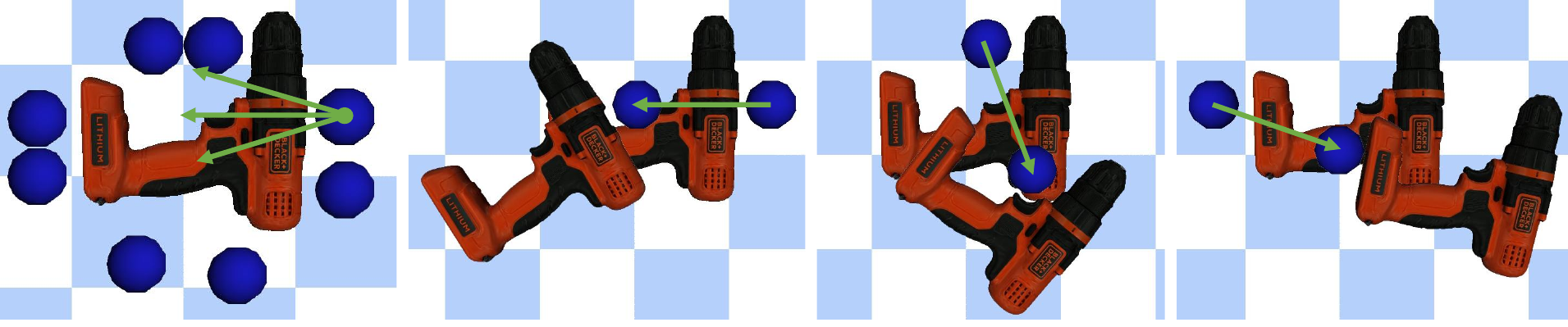}
    \caption{\revise{The pushing trajectories used in the experiments. Left: The 8 starting locations of the floating spherical robot pushing trajectories and 3 pushing directions towards the robot at one of the starting locations for the \texttt{Drill} object in the simulation experiments. Middle and right: the training trajectory and 2 sample testing trajectories, with the first and last frames shown. [Best viewed in color.]} }
    \label{fig:sim_trajectories}
\end{figure}

\begin{figure*}[t!]
\centering
    \includegraphics[trim=0cm 14.5cm 2cm 0cm,clip,width=0.9\linewidth]{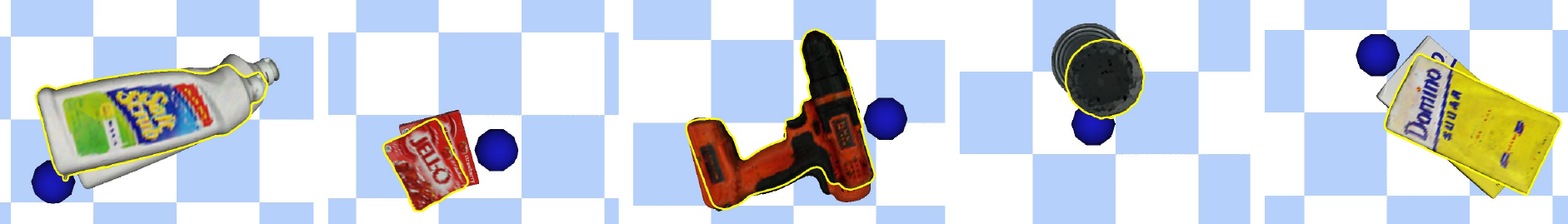}
    \caption{\revise{The predicted and ground-truth poses of the 5 different objects at the end of sampled testing trajectories for the simulation experiments. After training, the predicted poses are obtained by applying the control forces from the initial pose and integrating forward in time. The predicted object poses are highlighted with a yellow silhouette and overlaid with the ground-truth object, blue floating spherical robot, and background.} [Best viewed in color.]}
    \label{fig:sim_pushed}
\end{figure*}

\begin{figure*}[t!]
\centering
    \includegraphics[trim=0cm 15cm 0.5cm 0cm,clip,width=1\linewidth]{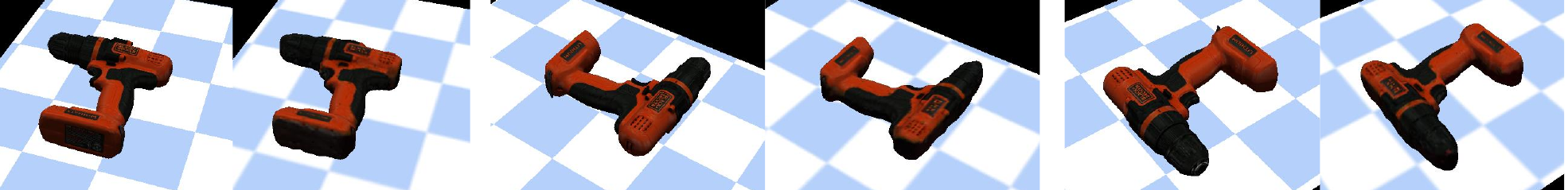}
    \put(-436,-8){\small View 1}
    \put(-267,-8){\small View 2 }
    \put(-96,-8){\small View 3}
    \caption{\revise{The ground-truth (left) and predicted (right) RGB images of 3 novel views of the \texttt{Drill} in simulation. The optimized mesh shape and geometry match the ground truth well, although lacking the fine details that can not be observed from the top view. The terrain checkers are not as sharp as the ground truth due to the use of point rendering of the colored terrain point cloud.}}
    \label{fig:sim_novelview}
\end{figure*}

\subsubsection{Two-stage Optimization} 
We note that while the initial guess can deviate significantly from our observations, deviations in geometry and appearance can be largely corrected by considering only the first observation, i.e. $\langle {t_1},e^{t_1},f^{t_1},o^{t_1},c^{t_1},m^{t_1}\rangle$. Therefore, our first stage considers only the first time instance and optimizes $\theta$ using the following loss:
\begin{align*}
\mathcal{L}(\hat{o}^{t_1},o^{t_1}) = 
    &c_1\mathcal{L}_\text{rgb}(\hat{o}^{t_1},o^{t_1}) + c_2\mathcal{L}_\text{depth}(\hat{o}^{t_1},o^{t_1}) + \\
    &c_3\mathcal{L}_\text{pcd}(o^{t_1},\theta) +  c_4\mathcal{L}_\text{pen}(\theta) + c_5 \mathcal{L}_\text{balance}(\theta)+\\
    &c_6\mathcal{L}_\text{reg}(\theta, \theta_0) + c_7\mathcal{L}_\text{smooth}(\mathcal{P}),
\end{align*}
with $(c_1,\dots,c_7)$ denoting weight terms. Here, $\hat{o}^{t_1}$ is the rendered RGB-D image at $t_1$. $\mathcal{L}_\text{rgb}$ is a loss on the RGB images, defined as a weighted sum of $l_1$ distance and D-SSIM terms: $\mathcal{L}_\text{rgb} = (1 - \lambda)\mathcal{L}_{1} + \lambda\mathcal{L}_\text{SSIM}$, where we set $\lambda = 0.2$. $\mathcal{L}_\text{depth}$ is the $l_1$ distance on the depth images. When calculating the three loss terms $\mathcal{L}_{1,\text{SSIM},\text{depth}}$, the robot is masked out according to the segmentation mask $m^{t_1}$. $\mathcal{L}_\text{pcd}$ is a unilateral Chamfer distance between the point cloud generated from the observed RGB-D image pixels belonging to the object and the mesh vertices generated from $\mathcal{P}$, which is defined as the average of the minimum distance between each point on the observed point cloud and any mesh vertex. The use of a unilateral Chamfer distance as opposed to the regular Chamfer distance is necessary since the observed object point cloud only includes points on the object surfaces visible from the camera viewpoint. The terms $\mathcal{L}_\text{pen}$ and $\mathcal{L}_\text{balance}$ encourage the object to achieve static force equilibrium while having no penetrations. $\mathcal{L}_\text{pen}$ penalizes the object mesh penetrations in the terrain, and $\mathcal{L}_\text{balance} = \sum_{1}^k\|p_o^{i+t_1}-p_o^{t_1}\|_1$ is the sum of the positional changes of the object from the initial position $p_o^{t_1}$ over $k$ steps by simulating forward with no robot actions. In the case where the initial guess of the object geometry does not come in contact with the terrain at $t_1$, $\mathcal{L}_\text{balance}$ allows the computation of the gradient information to expand the geometry towards the terrain once the object falls due to gravity and contacts the terrain during the $k$ steps. We use $k=3$ for all the experiments in this paper. These two terms provide a strong hint for the occluded part of the object. For example, when an object is lying on the table, our RGB-D observation will not cover the bottom of the object. However, our model will guide SaP to fill the bottom-side geometries by encouraging the object to settle on the table. Finally, the last two terms regularize the object shape, where ${L}_\text{reg}$ is the $L_2$ norm between the SaP points and the initial SaP points and $L_{smooth}$ is the Laplacian smoothing objective on the object mesh.

In addition, the use of both $\mathcal{L}_\text{depth}$ and $\mathcal{L}_\text{pcd}$ is necessary. When only $\mathcal{L}_\text{depth}$ is used, if the estimated mesh is smaller than the ground-truth object geometry, the predicted depth pixels that are supposed to reach the ground-truth mesh do not hit the estimated mesh, and there is no gradient information for expanding the geometry. Similarly, $\mathcal{L}_\text{pcd}$ does not inform the SaP to not expand over the observed point cloud, and $\mathcal{L}_\text{depth}$ prevents the object geometries from occupying the supposed background. 

After the first stage, we have tuned our model to match the first observation. When we move on to the second stage, we incorporate all timesteps by applying the robot controls. To make sure the geometry and appearance of the object stay close to that in the first time step, we still use the loss from the first stage on the first time frame, except for the penetration and balance losses. For the rest of the time frames, we use the loss $\mathcal{L}= c_8\mathcal{L}_\text{pcd} + c_{9}\mathcal{L}_\text{robot}$, where the first term has the same definition as in the first stage and $\mathcal{L}_\text{robot}$ is the squared distance between the ground-truth and predicted robot end-effector positions. To calculate this loss, we apply the robot control forces from the initial state, integrate forward in time, and render \revise{two intermediate and the last frames}, instead of every time frame for computational efficiency. During our optimization, the chain of gradients is back-propagated through the following recursive rule:
\begin{align*}
\FDD{\mathcal{L}(\hat{o}^{t_i},o^{t_i})}{\theta}=
\FPP{\mathcal{L}}{\theta}+
\FPP{\mathcal{L}}{q^{i}}
\left[\FPP{q^{i}}{\theta}+\FPP{q^{i}}{q^{i-1}}\FDD{q^{i-1}}{\theta}\right],
\end{align*}
where the first two terms $\FPPROW{\mathcal{L}}{\theta}, \FPPROW{\mathcal{L}}{q^i}$ is the derivatives of the rendering equation, and the remaining terms in the bracket are the derivatives of the simulator.

\begin{table*}[]
    \renewcommand{\arraystretch}{1}
    \centering
    \footnotesize
    \begin{tabular}{@{}c|cc|cccc@{}}
        \toprule
         & \multicolumn{2}{c|}{Dynamics Parameter Error} & \multicolumn{4}{c}{Trajectory Prediction Error}\\
         \midrule
         Method & mass (kg) & $\mu$ & Unilateral Chamfer\,(mm) & Pos.\,(mm) & Rot.\,($^\circ$) & Trans. Vel. (m/s$^2$) \\
         \midrule
        Ours       & 0.0728  & 0.106 &  8.69          & 15.5      & 16.7      &   0.0351   \\
        PhysGS~\cite{abou-chakra2024physically} &   0.225      &    0.400   & 24.2 & 42.8 & 31.8 & 0.436  \\
        \bottomrule
    \end{tabular}
    \caption{\revise{Average Dynamics Parameter Identification and novel trajectory prediction errors for all objects in the simulation experiments}}
    \vspace{-10px}
    \label{tab:results}
\end{table*} 

\subsubsection{Geometry Prior}
Our method relies on a reasonable initial guess. Imagine the case with an object settling on the edge of a table and the camera does not observe the contact between the two. The initial guess of the occluded part of the object could be very short and cause the object to directly fall down without touching the table. This cannot be recovered by our optimization since the object never hits the table and there are no gradients for correcting the geometries. To obtain a reasonable initial guess of the geometries and appearance of the rigid object of interest from partial visual observations, we take advantage of large reconstruction models~\cite{hong2023lrm} that predict object 3D models from a single RGB image, trained on web-scale data. In particular, we use TripoSR~\cite{TripoSR2024} in our experiments with the segmented RGB image of the object as the input image. Since the generated mesh is scale- and transform-agnostic, we apply RANSAC and the scale-aware iterative closest point algorithms with Open3D~\cite{Zhou2018} to register the mesh to the partial object point cloud, computed from the RGB-D image at the first time instance.

Finally, in all the experiments of this work, we assume that the occluded terrain by the object is flat, and complete the terrain by fitting a plane of points, where the colors match the nearest visible points of the terrain. In addition, in all the experiments, we do not optimize the point cloud position of the terrain and optimize only the colors. Although these settings are simplifying, we believe a similar approach could be adopted that predicts the geometry of the occluded rigid terrain from a geometry prior model and optimizes for the terrain geometry simultaneously, although more online data may be required to resolve the ambiguities of the contacts between two occluded geometries.

\section{Experiments and Results}\label{sec:result}
To validate our method, we first conduct experiments with simulated data, and then in the real world.


\begin{figure}[]
\centering
    \includegraphics[trim=0cm 16.2cm 14.8cm 0cm,clip,width=1\linewidth]{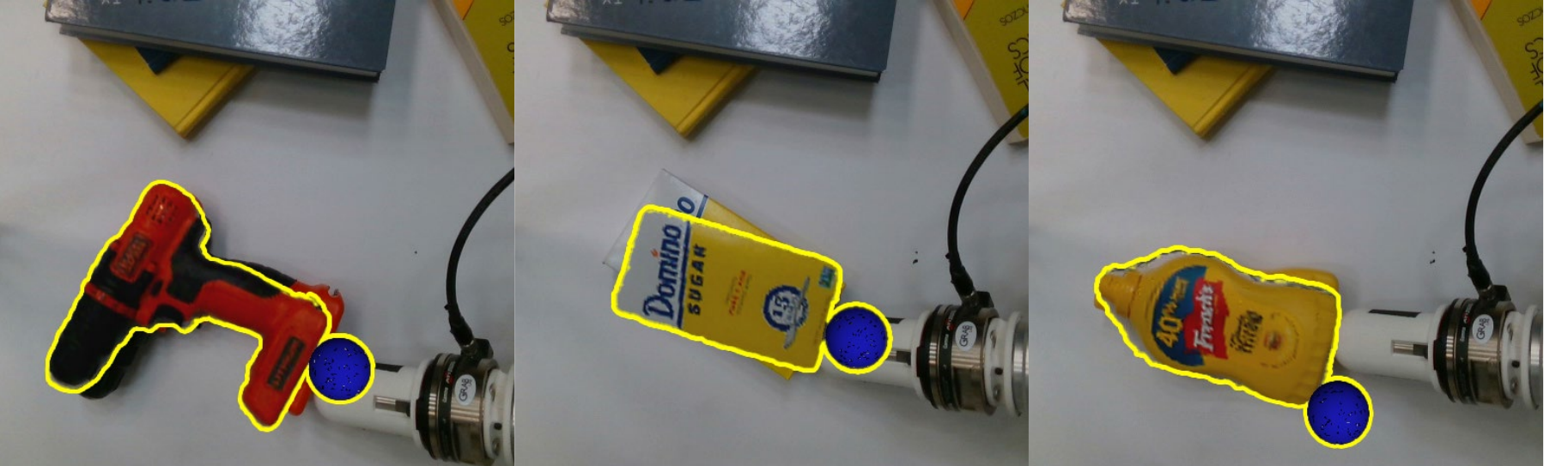}
    \caption{\revise{Results of three example testing trajectory of the physical experiments. The predicted object and robot poses with the optimized $\theta$ highlighted with a yellow silhouette are overlaid with the ground-truth object, robot, and background.} [Best viewed in color.]}
    \label{fig:real_results}
\end{figure}

\begin{figure}[]
\centering
    \includegraphics[trim=0cm 12.5cm 1.2cm 0cm,clip,width=1\linewidth]{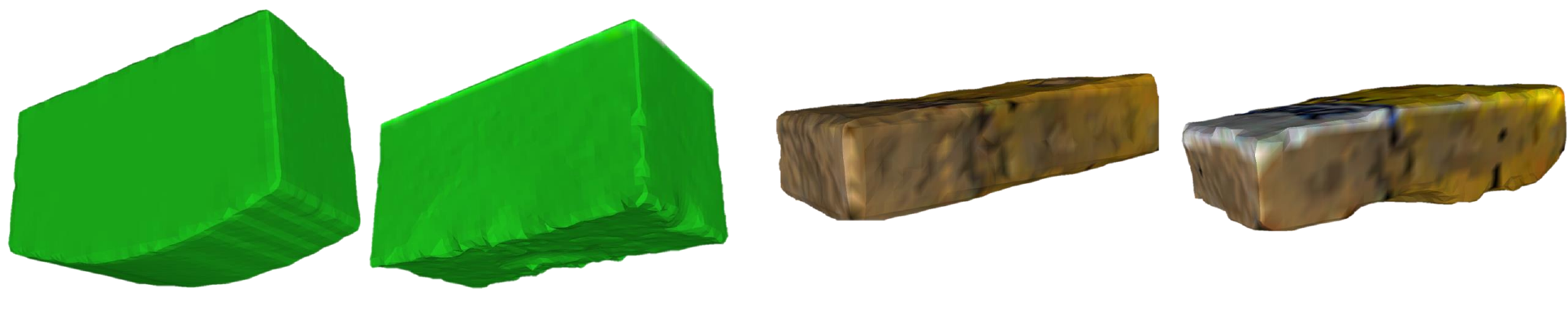}
    \put(-228,-7){\small Initial}
    \put(-180,-7){\small Optimized}
    \put(-105,-7){\small Initial}
    \put(-50,-7){\small Optimized}
    \caption{The initial guess and optimized shape of the \texttt{Box} object in simulated experiments and the \texttt{Sugar} object in physical experiments during stage 1. The algorithm is able to correct the bulging on the underside of the \texttt{Box} that would intersect the terrain. On the other hand, the initial mesh of \texttt{Sugar} is too thin and does not touch the terrain. Our algorithm is able to optimize the shape so that it satisfies physics constraints.}
    \label{fig:block_optimized}
    \vspace{-10px}
\end{figure}

\subsection{Simulation Experiments}
Shown in Fig.~\ref{fig:setups}, we conduct all the simulated experiments using data collected with the PyBullet simulator~\cite{coumans2021}. \revise{We use a simple green box object (\texttt{Box}) and 8 objects (\texttt{Gelatin}, \texttt{RubiksCube}, \texttt{Spam}, \texttt{TunaCan}, \texttt{Mustard}, \texttt{Bleach}, \texttt{Drill}, and \texttt{Sugar}) from the YCB object dataset~\cite{7254318}, which covers diverse shapes, sizes, and textures.} The objects are placed on a flat surface with checker patterns and pushed by a floating sphere robot, while a static overhead camera takes pictures. \revise{As shown in Fig.~\ref{fig:sim_trajectories}, 24 pushing trajectories are adopted, where one is used to optimize the world model and the rest for evaluating the optimized model. To make sure that the pushes are diverse, we pick 2 starting locations on each of the four sides of the object, and push in three directions that are 20$^\circ$ apart at each of these 8 locations toward the object. Similar pushes are used for all the other objects where the starting locations are adjusted based on the size of the objects. The trajectories push the objects up to 12\,cm and 80$^\circ$}. All the trajectories have $T=30$ time steps with $\delta t = 0.01$\,s. We use the following weights for optimization: $[c_1,c_2,c_3,c_4,c_5,c_6,c_7,c_8,c_9] = [10,500,2000,100,100,100,4000,500,100]$. These terms are not carefully tuned and are set such that each term has a similar order of magnitude at the start of optimization for our experiments. \revise{The physical parameters we optimize for include the mass, surface coefficient of friction, the center of mass, and the rotational inertial. The center of mass is initialized at the geometry center of the initial shape guess. The rotational inertia is initialized by treating the object as a box, whose dimensions are the bounding box of the initial geometry guess. We assume that the rotational inertia only has diagonal terms. For all experiments including simulations and real-world experiments, the surface coefficient of friction is initialized at 0.2 and the mass is initialized at 0.2\,kg.} 

\revise{We first evaluate whether our method can identify the shape and physical parameters accurately, such that it generalizes well to new physical interactions. We compare our method against a recently proposed method that represents the world jointly with Gaussian splats and physical particles and allows it to perform both novel-view rendering and physics-based trajectory predictions~\cite{abou-chakra2024physically}. We refer to this method as PhysGS. This method performs simulation with a particle-based simulator~\cite{warp2022}. In the original paper, the physical parameters of the particles are arbitrarily set, and virtual forces are optimized to match the observations and predictions on the training trajectory. To allow more accurate prediction for new trajectories, we optimize the total mass and coefficient of friction of the object particles with grid search using the partial Chamfer distance between the observed object point cloud and the predicted object particles on the last time frame in the trajectory.} 

\revise{We report the quantitative results for dynamics parameter estimation and novel trajectory predictions for all objects in the simulation experiments in Table~\ref{tab:results}, which are the average across all objects. The dynamics parameter estimation error from the training trajectory, the average pose error, the unilateral Chamfer distance, and the translational velocity error at the end of the testing trajectories are reported. We also show some qualitative examples that are representative of the average errors in Fig.~\ref{fig:sim_pushed}. The final pose and velocity of the objects are obtained by applying the control forces and integrating forward in time. Our method identifies the dynamics parameters accurately and shows low trajectory prediction errors. We would like to point out in particular that the average rotation error is heavily skewed by the \texttt{TunaCan} object since our method currently uses the object surface point cloud to track the object during stage 2 optimization, and cannot properly differentiate the rotation of a cylindrical object. As a result, the average rotation error is 28.0$^\circ$ for \texttt{TunaCan}, and we aim to address this issue in future work. On the other hand, PhysGS generalizes very poorly to the testing trajectories. While this is partially because of the lack of proper dynamics parameters and physics-based shape estimation, we also find that the particle-based simulator is extremely sensitive to simulator parameters and have poor physical fidelity, especially for rigid objects. We also show an example of the initial and optimized geometries of the \texttt{Box} object in Fig.~\ref{fig:block_optimized} (left), which demonstrate the ability of our method to adjust occluded geometry based on the physics.} \revise{Next, we also evaluate the quality of the novel-view synthesis of our method. For each of the testing objects, we evaluate the synthesized RGB images from 10 novel viewpoints around the scene, and our method achieves 0.00225 of mean squared error (MSE), 0.965 of structural similarity index measure (SSIM) and 26.5 of peak signal-to-noise ratio (PSNR). We also show some examples of novel view synthesis of the \texttt{Drill} object in Fig.~\ref{fig:sim_novelview}, which matches the ground truth very well.}

\subsection{Physical Experiments}
Shown in Fig.~\ref{fig:setups}, we conduct physical experiments with a UR5e robot arm equipped with an ATI Gamma F/T sensor and a pusher with a semispherical end, and a static overhead RealSense D435 RGB-D camera. We use similar pushing trajectories to those for the simulation experiments, but only use 6 trajectories with two different starting locations from one side of the object. We then use one trajectory for training and the rest for evaluation with 3 YCB objects: power drill (\texttt{Drill}), sugar box (\texttt{Sugar}), and mustard bottle (\texttt{Mustard}). We use a total of $T=48$ time steps with $\delta t = 0.03$\,s. We use the same optimization settings as the simulated experiments.


On average across all testing trajectories and testing objects, our method achieves a mass identification error of 0.186\,kg and 6.10\,mm of unilateral chamfer distance between the observed object point cloud and the predicted object at the last frame of the trajectory. Note that we only reported the mass identification error because the ground truth is very easy to measure while measuring the surface friction requires a specialized setup. Since we do not have access to the ground-truth object poses and only have access to the raw observations, we report the unilateral Chamfer distances between the observed object point cloud and the predicted object geometry. The train and test prediction results for three sample trajectories are visualized in Fig.~\ref{fig:real_results}. Overall, the errors are comparable to those from the simulated experiments. In addition, we show the initial and optimized shape of the \texttt{Sugar} object in Fig.~\ref{fig:block_optimized} on the right, where the initial shape is too thin and does not contact the terrain below. Our algorithm modifies the occluded geometry to satisfy the physics. 


\section{Limitations}
Our method assumes ground-truth object masks from the scene, which might not always be possible even with advanced foundational segmentation models. In addition, our method does not consider more advanced appearance models, lighting sources, and shadows. As a result, the rendered scenes could have artifacts that do not match the real-world observations. Currently, each optimization run is completed in under 15\,mins on a standard PC with an Intel i9-13900KF CPU, 64 GB of RAM, and a GeForce RTX 4090 GPU. While this is not ideal for online robotics applications, we intend to reduce the runtime by using better initial guesses of geometry and physical parameters from data-driven pre-trained models and more efficient implementation. Finally, our method currently struggles on objects whose rotation can not be properly identified from a surface point cloud, such as a cylinder. We aim to explore tracking methods that also leverages surface textures for pose tracking in future work. 

\section{Conclusion}
We propose a novel algorithm to solve the task of identifying objects' physical properties as well as the geometry and appearance, a crucial step in downstream robot manipulation tasks. To the best of our knowledge, this is the first method that allows the joint optimization of all of these properties. Our method combines the merit of SaP object representation~\cite{liao2018deep}, differentiable collision detection~\cite{guendelman2003nonconvex}, and differentiable simulation~\cite{strecke2021_diffsdfsim}. Although our method has several limitations, it opens doors to a rich spectrum of future research topics. Some potential future directions include extending our method to identify multi-body dynamic systems with additional constraints, more advanced appearance models, and physics-based perception to correct for wrong object masks.
\FloatBarrier
\bibliographystyle{IEEEtran}
\bibliography{references}

\begin{thebibliography}{10}
\providecommand{\url}[1]{#1}
\csname url@samestyle\endcsname
\providecommand{\newblock}{\relax}
\providecommand{\bibinfo}[2]{#2}
\providecommand{\BIBentrySTDinterwordspacing}{\spaceskip=0pt\relax}
\providecommand{\BIBentryALTinterwordstretchfactor}{4}
\providecommand{\BIBentryALTinterwordspacing}{\spaceskip=\fontdimen2\font plus
\BIBentryALTinterwordstretchfactor\fontdimen3\font minus \fontdimen4\font\relax}
\providecommand{\BIBforeignlanguage}[2]{{%
\expandafter\ifx\csname l@#1\endcsname\relax
\typeout{** WARNING: IEEEtran.bst: No hyphenation pattern has been}%
\typeout{** loaded for the language `#1'. Using the pattern for}%
\typeout{** the default language instead.}%
\else
\language=\csname l@#1\endcsname
\fi
#2}}
\providecommand{\BIBdecl}{\relax}
\BIBdecl

\bibitem{ha2018world}
D.~Ha and J.~Schmidhuber, ``World models,'' \emph{arXiv preprint arXiv:1803.10122}, 2018.

\bibitem{Bruce2024world}
J.~Bruce, M.~D. Dennis, A.~Edwards, J.~Parker-Holder, Y.~Shi, E.~Hughes, M.~Lai, A.~Mavalankar, R.~Steigerwald, C.~Apps \emph{et~al.}, ``Genie: Generative interactive environments,'' 2024.

\bibitem{Zhu2024world}
F.~Zhu, H.~Wu, S.~Guo, Y.~Liu, C.~Cheang, and T.~Kong, ``Irasim: Learning interactive real-robot action simulators,'' \emph{arXiv preprint arXiv:2406.14540}, 2024.

\bibitem{Yang2023world}
M.~Yang, Y.~Du, K.~Ghasemipour, J.~Tompson, L.~Kaelbling, D.~Schuurmans, P.~Abbeel, U.~Berkeley, and G.~DeepMind, ``Learning interactive real-world simulators,'' \emph{International Conference on Learning Representations}, 2024.

\bibitem{Song-RSS-20}
C.~Song and A.~Boularias, ``{Learning to Slide Unknown Objects with Differentiable Physics Simulations},'' in \emph{Proceedings of Robotics: Science and Systems}, Corvalis, Oregon, USA, July 2020.

\bibitem{9363565}
Q.~Le~Lidec, I.~Kalevatykh, I.~Laptev, C.~Schmid, and J.~Carpentier, ``Differentiable simulation for physical system identification,'' \emph{IEEE Robotics and Automation Letters}, vol.~6, no.~2, pp. 3413--3420, 2021.

\bibitem{ehsani2020use}
K.~Ehsani, S.~Tulsiani, S.~Gupta, A.~Farhadi, and A.~Gupta, ``Use the force, luke! learning to predict physical forces by simulating effects,'' \emph{IEEE Conference on Computer Vision and Pattern Recognition}, 2020.

\bibitem{mildenhall2021nerf}
B.~Mildenhall, P.~P. Srinivasan, M.~Tancik, J.~T. Barron, R.~Ramamoorthi, and R.~Ng, ``Nerf: Representing scenes as neural radiance fields for view synthesis,'' \emph{Communications of the ACM}, vol.~65, no.~1, pp. 99--106, 2021.

\bibitem{kerbl20233d}
B.~Kerbl, G.~Kopanas, T.~Leimk{\"u}hler, and G.~Drettakis, ``3d gaussian splatting for real-time radiance field rendering.'' \emph{ACM Trans. Graph.}, vol.~42, no.~4, pp. 139--1, 2023.

\bibitem{erez2015simulation}
T.~Erez, Y.~Tassa, and E.~Todorov, ``Simulation tools for model-based robotics: Comparison of bullet, havok, mujoco, ode and physx,'' in \emph{IEEE international conference on robotics and automation}, 2015.

\bibitem{peng2021sap}
S.~Peng, C.~Jiang, Y.~Liao, M.~Niemeyer, M.~Pollefeys, and A.~Geiger, ``Shape as points: A differentiable poisson solver,'' \emph{Advances in Neural Information Processing Systems}, vol.~34, pp. 13\,032--13\,044, 2021.

\bibitem{liao2018deep}
Y.~Liao, S.~Donne, and A.~Geiger, ``Deep marching cubes: Learning explicit surface representations,'' in \emph{IEEE conference on computer vision and pattern recognition}, 2018.

\bibitem{strecke2021_diffsdfsim}
M.~Strecke and J.~Stueckler, ``{DiffSDFSim}: Differentiable rigid-body dynamics with implicit shapes,'' in \emph{International Conference on {3D} Vision ({3DV})}, Dec. 2021.

\bibitem{Werling-RSS-21}
K.~Werling, D.~Omens, J.~Lee, I.~Exarchos, and C.~K. Liu, ``{Fast and Feature-Complete Differentiable Physics Engine for Articulated Rigid Bodies with Contact Constraints},'' in \emph{Proceedings of Robotics: Science and Systems}, Virtual, July 2021.

\bibitem{Xu-RSS-21}
J.~Xu, T.~Chen, L.~Zlokapa, M.~Foshey, W.~Matusik, S.~Sueda, and P.~Agrawal, ``{An End-to-End Differentiable Framework for Contact-Aware Robot Design},'' in \emph{Proceedings of Robotics: Science and Systems}, Virtual, July 2021.

\bibitem{NEURIPS2021_8e296a06}
Y.~Qiao, J.~Liang, V.~Koltun, and M.~Lin, ``Differentiable simulation of soft multi-body systems,'' \emph{Advances in Neural Information Processing Systems}, 2021.

\bibitem{howell2022dojo}
T.~A. Howell, S.~L. Cleac'h, J.~Br{\"u}digam, J.~Z. Kolter, M.~Schwager, and Z.~Manchester, ``Dojo: A differentiable physics engine for robotics,'' \emph{arXiv preprint arXiv:2203.00806}, 2022.

\bibitem{geilinger2020add}
M.~Geilinger, D.~Hahn, J.~Zehnder, M.~B{\"a}cher, B.~Thomaszewski, and S.~Coros, ``Add: Analytically differentiable dynamics for multi-body systems with frictional contact,'' \emph{ACM Transactions on Graphics (TOG)}, vol.~39, no.~6, pp. 1--15, 2020.

\bibitem{achlioptas2018learning}
P.~Achlioptas, O.~Diamanti, I.~Mitliagkas, and L.~Guibas, ``Learning representations and generative models for 3d point clouds,'' in \emph{International conference on machine learning}.\hskip 1em plus 0.5em minus 0.4em\relax PMLR, 2018, pp. 40--49.

\bibitem{deng2020cvxnet}
B.~Deng, K.~Genova, S.~Yazdani, S.~Bouaziz, G.~Hinton, and A.~Tagliasacchi, ``Cvxnet: Learnable convex decomposition,'' in \emph{Proceedings of the IEEE/CVF conference on computer vision and pattern recognition}, 2020, pp. 31--44.

\bibitem{pfrommer2021contactnets}
S.~Pfrommer, M.~Halm, and M.~Posa, ``Contactnets: Learning discontinuous contact dynamics with smooth, implicit representations,'' in \emph{Conference on Robot Learning}.\hskip 1em plus 0.5em minus 0.4em\relax PMLR, 2021, pp. 2279--2291.

\bibitem{sharp2022spelunking}
N.~Sharp and A.~Jacobson, ``Spelunking the deep: Guaranteed queries on general neural implicit surfaces via range analysis,'' \emph{ACM Transactions on Graphics (TOG)}, vol.~41, no.~4, pp. 1--16, 2022.

\bibitem{xie2024physgaussian}
T.~Xie, Z.~Zong, Y.~Qiu, X.~Li, Y.~Feng, Y.~Yang, and C.~Jiang, ``Physgaussian: Physics-integrated 3d gaussians for generative dynamics,'' in \emph{Proceedings of the IEEE/CVF Conference on Computer Vision and Pattern Recognition}, 2024, pp. 4389--4398.

\bibitem{aastrom1971system}
K.~J. {\AA}str{\"o}m and P.~Eykhoff, ``System identification—a survey,'' \emph{Automatica}, vol.~7, no.~2, pp. 123--162, 1971.

\bibitem{abou-chakra2024physically}
J.~Abou-Chakra, K.~Rana, F.~Dayoub, and N.~Suenderhauf, ``Physically embodied gaussian splatting: A visually learnt and physically grounded 3d representation for robotics,'' in \emph{8th Annual Conference on Robot Learning}, 2024.

\bibitem{kazhdan2006poisson}
M.~Kazhdan, M.~Bolitho, and H.~Hoppe, ``Poisson surface reconstruction,'' in \emph{Proceedings of the fourth Eurographics symposium on Geometry processing}, vol.~7, no.~4, 2006.

\bibitem{liu2019soft}
S.~Liu, T.~Li, W.~Chen, and H.~Li, ``Soft rasterizer: A differentiable renderer for image-based 3d reasoning,'' in \emph{Proceedings of the IEEE/CVF international conference on computer vision}, 2019, pp. 7708--7717.

\bibitem{Laine2020diffrast}
S.~Laine, J.~Hellsten, T.~Karras, Y.~Seol, J.~Lehtinen, and T.~Aila, ``Modular primitives for high-performance differentiable rendering,'' \emph{ACM Transactions on Graphics}, vol.~39, no.~6, 2020.

\bibitem{ravi2020pytorch3d}
N.~Ravi, J.~Reizenstein, D.~Novotny, T.~Gordon, W.-Y. Lo, J.~Johnson, and G.~Gkioxari, ``Accelerating 3d deep learning with pytorch3d,'' \emph{arXiv:2007.08501}, 2020.

\bibitem{wiles2020synsin}
O.~Wiles, G.~Gkioxari, R.~Szeliski, and J.~Johnson, ``Synsin: End-to-end view synthesis from a single image,'' in \emph{Proceedings of the IEEE/CVF conference on computer vision and pattern recognition}, 2020, pp. 7467--7477.

\bibitem{anitescu1997formulating}
M.~Anitescu and F.~A. Potra, ``Formulating dynamic multi-rigid-body contact problems with friction as solvable linear complementarity problems,'' \emph{Nonlinear Dynamics}, vol.~14, pp. 231--247, 1997.

\bibitem{cline2002rigid}
M.~B. Cline, ``Rigid body simulation with contact and constraints,'' Ph.D. dissertation, University of British Columbia, 2002.

\bibitem{stewart1996implicit}
D.~E. Stewart and J.~C. Trinkle, ``An implicit time-stepping scheme for rigid body dynamics with inelastic collisions and coulomb friction,'' \emph{International Journal for Numerical Methods in Engineering}, vol.~39, no.~15, pp. 2673--2691, 1996.

\bibitem{hong2023lrm}
Y.~Hong, K.~Zhang, J.~Gu, S.~Bi, Y.~Zhou, D.~Liu, F.~Liu, K.~Sunkavalli, T.~Bui, and H.~Tan, ``Lrm: Large reconstruction model for single image to 3d,'' \emph{arXiv preprint arXiv:2311.04400}, 2023.

\bibitem{TripoSR2024}
D.~Tochilkin, D.~Pankratz, Z.~Liu, Z.~Huang, , A.~Letts, Y.~Li, D.~Liang, C.~Laforte, V.~Jampani, and Y.-P. Cao, ``Triposr: Fast 3d object reconstruction from a single image,'' \emph{arXiv:2403.02151}, 2024.

\bibitem{Zhou2018}
Q.-Y. Zhou, J.~Park, and V.~Koltun, ``{Open3D}: {A} modern library for {3D} data processing,'' \emph{arXiv:1801.09847}, 2018.

\bibitem{coumans2021}
E.~Coumans and Y.~Bai, ``Pybullet, a python module for physics simulation for games, robotics and machine learning,'' \url{http://pybullet.org}, 2016--2021.

\bibitem{7254318}
B.~Calli, A.~Walsman, A.~Singh, S.~Srinivasa, P.~Abbeel, and A.~M. Dollar, ``Benchmarking in manipulation research: Using the yale-cmu-berkeley object and model set,'' \emph{IEEE Robotics and Automation Magazine}, vol.~22, no.~3, pp. 36--52, 2015.

\bibitem{warp2022}
M.~Macklin, ``Warp: A high-performance python framework for gpu simulation and graphics,'' 2022.

\bibitem{guendelman2003nonconvex}
E.~Guendelman, R.~Bridson, and R.~Fedkiw, ``Nonconvex rigid bodies with stacking,'' \emph{ACM transactions on graphics (TOG)}, vol.~22, no.~3, pp. 871--878, 2003.

\end{thebibliography}
\end{document}